\documentclass[10pt,twocolumn,letterpaper]{article}

\usepackage{iccv}
\usepackage{times}
\usepackage{epsfig}
\usepackage{graphicx}
\usepackage{amsmath}
\usepackage{amssymb}
\usepackage{booktabs}
\usepackage{algorithm}
\usepackage{algorithmic}
\usepackage{setspace}
\usepackage[accsupp]{axessibility}

\usepackage[breaklinks=true,bookmarks=false]{hyperref}
\hypersetup{colorlinks=true}

\iccvfinalcopy 

\def\iccvPaperID{****} 

\ificcvfinal\fi

\begin{document}

\title{Cross Contrasting Feature Perturbation for Domain Generalization}

\author{Chenming Li$^1$  \quad  Daoan Zhang$^1$ \quad
Wenjian Huang$^1$ \quad
Jianguo Zhang$^{1,2,}$\thanks{Corresponding author.}\\
$^1$Research Institute of Trustworthy Autonomous Systems and Department of Computer Science \\ and Engineering, Southern University of Science and Technology, Shenzhen, China\\
$^2$Peng Cheng Laboratory, Shenzhen, China\\
{\tt\small 12132339@mail.sustech.edu.cn, 12032503@mail.sustech.edu.cn, \{huangwj, zhangjg\}@sustech.edu.cn,}\\
}


\maketitle
\ificcvfinal\thispagestyle{empty}\fi

\begin{abstract}
   Domain generalization (DG) aims to learn a robust model from source domains that generalize well on unseen target domains. Recent studies focus on generating novel domain samples or features to diversify distributions complementary to source domains. Yet, these approaches can hardly deal with the restriction that the samples synthesized from various domains can cause semantic distortion. In this paper, we propose an online one-stage \underline{C}ross \underline{C}ontrasting \underline{F}eature \underline{P}erturbation (CCFP) framework to simulate domain shift by generating perturbed features in the latent space while regularizing the model prediction against domain shift. Different from the previous fixed synthesizing strategy, we design modules with learnable feature perturbations and semantic consistency constraints. In contrast to prior work, our method does not use any generative-based models or domain labels. We conduct extensive experiments on a standard DomainBed benchmark with a strict evaluation protocol for a fair comparison. Comprehensive experiments show that our method outperforms the previous state-of-the-art, and quantitative analyses illustrate that our approach can alleviate the domain shift problem in out-of-distribution (OOD) scenarios. \href{https://github.com/hackmebroo/Cross-Contrasting-Feature-Perturbation-for-Domain-Generalization/tree/main}{https://github.com/hackmebroo/CCFP}
\end{abstract}

\section{Introduction}
\label{sec:intro}

Deep Neural Networks have achieved remarkable success on a number of computer vision tasks\cite{lecun2015deep, zhang2023rethinking}. These models rely on the $i.i.d$ assumption\cite{vapnik1991principles}, $i.e.$, the training data and testing data are identically and independently distributed. However, in real-world scenarios, the assumption does not always hold due to the $domain\;shift$ problem\cite{ben2010theory}. For instance, it is hard for a model trained on photographs to adapt to sketches. 

Domain adaptation (DA) methods\cite{ganin2015unsupervised,sun2016deep,wilson2020survey} can be employed to handle the out-of-distribution (OOD) issue in the settings where unlabeled target data is available. Although DA can perform well on known target domains, it still fails in practical situations where target domains are not accessible during training. Domain generalization (DG) \cite{wang2022generalizing} aims to deal with such problems. The goal of domain generalization is to learn a generalized model from multiple different but related source domains ($i.e.$ diverse training datasets with the same label space) that can perform well on arbitrary unseen target domains. To realize this goal, most deep learning models are trained to minimize the average loss over the training set, which is known as the Empirical Risk Minimization (ERM) principle\cite{vapnik1999nature}. However, ERM-based network provably fails to OOD scenarios\cite{nagarajan2020understanding,eastwood2022probable, huang2022density, zhang2023aggregation}. 



One line of work\cite{sinha2018certifying,robey2021model,sagawa2019distributionally} improves the generalization capability of a model by optimizing the worst-domain risk over the set of possible domains, which are created by perturbing samples in the image level or using generative-based model ($i.e.$ VAE\cite{kingma2014auto} or GAN\cite{goodfellow2014generative}) to generate fictitious samples. Despite the performance promoted by creating samples in the image level on an offline basis to approximate the $worst$ $case$ over the entire family of domains, it is hard to generate "fictitious" samples in the input space without losing semantic discriminative information\cite{qiao2020learning}. Moreover, the offline two-stage data perturbation training procedure is nontrivial since both training a generative-based model and inferring them to obtain perturbed samples are challenging tasks. 

Another line of work perturbs features in the latent space\cite{verma2019manifold,zhou2020domain} by tuning the scaling and shifting parameters after instance normalization. Another study\cite{li2021uncertainty} extends it and leverages the uncertainty associated with feature statistics perturbation.
However, these methods all rely on a fixed perturbation strategy (linear interpolation or random perturbation) which limits the domain transportation from synthesized features to original features. Besides, although the instance normalization-based feature perturbation can change the information of intermediate features which is specific to domains, they still fail to preserve the semantic invariant, as the instance normalization may dilute discriminative information that is relevant to task objectives\cite{nam2018batch}. The performance of feature synthesis methods can be undetermined on account of semantic inconsistency\cite{lu2022semantic}. 

As is mentioned above, the data perturbation based methods can hardly generate the fictitious samples in the input space, and the feature perturbation based methods limits the diversity of the synthesized features and fail to preserve the semantic consistency. To address both of these issues, we propose to enforce a domain-aware adaptive feature perturbation in the latent space following the worst-case optimization objective and explicitly constrain the semantic consistency to preserve the class discriminative information.

\begin{figure*}[t]
\centering
\includegraphics[width=1\textwidth]{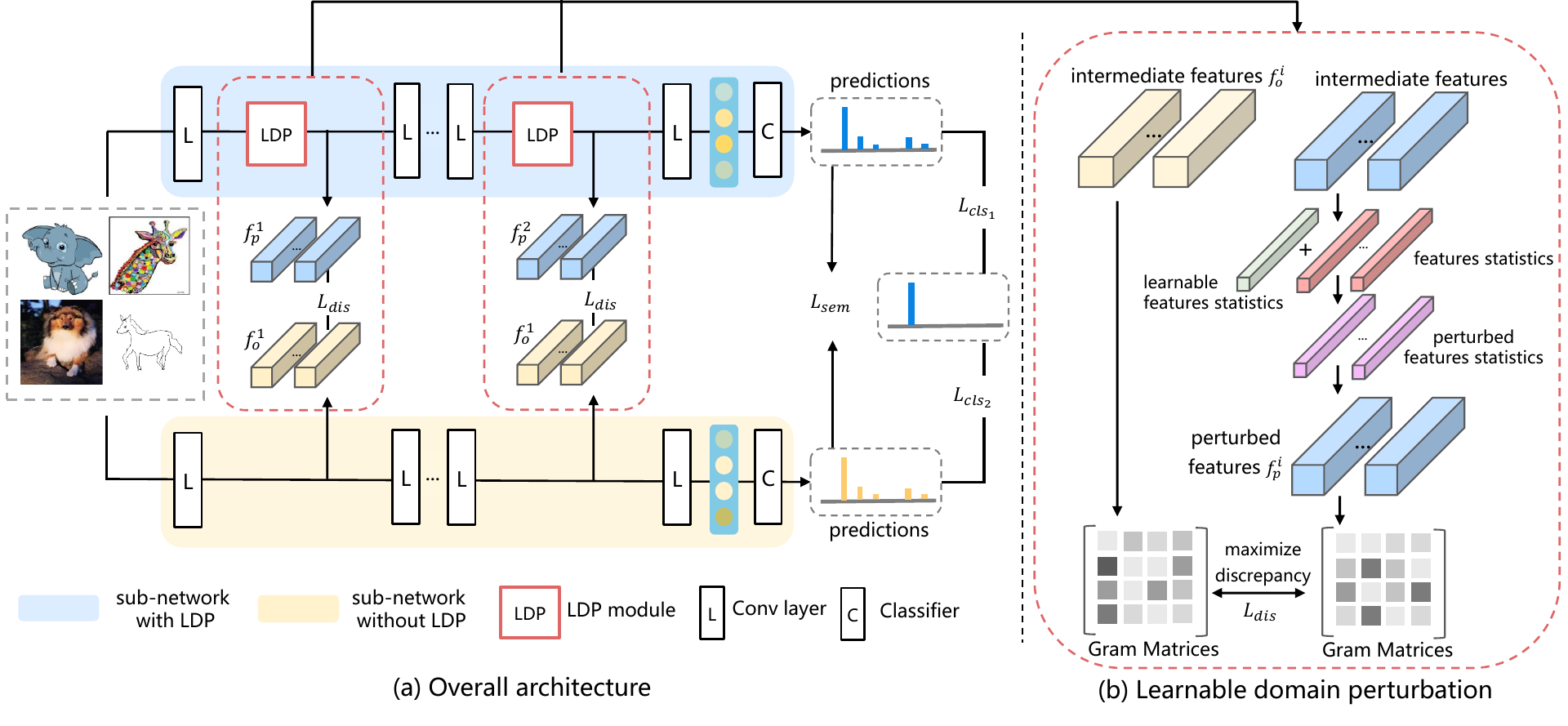} 
\caption{{\bf An overview of our proposed CCFP.} Our framework consists of two sub-networks. (a) The bottom network is a pre-trained backbone, and the top network is the same pre-trained backbone equipped with LDP modules (red boxes). The two sub-networks have similar architecture (except for the LDP modules) but do not share parameters. The steps of feature perturbation and the calculation of $\mathcal{L}_{dis}$ are shown in (b). 
}
\label{main}
\end{figure*}

Practically, the desideratum for the worst-case DG problem is to simulate the realistic domain shift by maximizing the domain discrepancy and minimizing the class discriminative characteristics between the source domain distribution and the fictitious target domain distribution. To this end, we design an adaptive online one-stage $Cross$ $Contrasting$ $Feature$ $Perturbation$ (CCFP) framework. An illustration of CCFP is shown in Figure \ref{main}. Our CCFP consists of two sub-network, one is used to extract the original features which represent the online estimate of the source distribution, and the other is used to perturb features in the latent space to create semantic invariant fictitious target distribution. In order to preserve the class discriminative information of the perturbed features, we regularize the predictions between the two sub-networks.

A key component of our framework is the feature perturbation. As pointed out in the research field of style transfer\cite{dumoulin2016learned,huang2017arbitrary}, the feature statistics carry the information primarily referring to domain-specific but
are less relevant to class discriminative. Based on this, we design a $learnable$ $domain$ $perturbation$ (LDP) module which can generate learnable perturbation of features to enlarge the domain transportation from the original ones. Note that the LDP only adds learnable scaling and shifting parameters on feature statistics without adopting domain labels or additional generative models. 

Another critical point of CCFP is the measurement of domain discrepancy. Different from existing study measure the domain discrepancy in the last layer\cite{wang2018visual, tzeng2014deep}, we propose to measure the domain discrepancy from the intermediate features to align with the observation that the shallow layers of the network learn low-level features (such as color and edges) which are more domain aware but less semantic relevant\cite{wang2019implicit}. Additionally, Gatys et al.\cite{gatys2015texture} show that Gram matrices of latent features can be used to encode stylistic attributes like textures and patterns. Motivated by this, we  develop a novel Gram-matrices-based metric to represent the domain-specific information from the intermediate activations. We maximize the dissimilarity between the intermediate features' Gram matrices to simulate the domain shift.  

We validate the effectiveness of CCFP on a standard DG benchmark called Domainbed\cite{gulrajani2020search}. Comprehensive experiment results show that our method surpasses previous methods and achieves state-of-the-art.

 In summary, our contributions are three-fold:
\begin{itemize}
    \item We propose a novel online one-stage cross contrasting feature perturbation framework (CCFP) for worst-case domain generalization problem, which can generate perturbed features while regularizing semantic consistency.
    \item We develop a learnable domain perturbation (LDP) module and an effective domain-aware Gram-matrices-based metric to measure domain discrepancy, which are useful for DG and integrated into the above CCFP framework. Additionally, our algorithm does not use any generative-based models and domain labels.
    \item Comprehensive experiments show that our method achieves state-of-the-art performance on diverse DG benchmarks under strict evaluation protocols of DomainBed\cite{gulrajani2020search}.
    
\end{itemize}

\section{Related Work}

\textbf{Domain generalization (DG)} aims to learn generalized representations from multiple source domains that can generalize well on arbitrary unseen target domains. For example, in the PACS dataset\cite{li2017deeper}, the task is to extract category-related knowledge, and the domains correspond to different artistic styles like art-painting, cartoon, photo, and sketch. The model will use three of four datasets to train and use the rest dataset to test. Various methods have been proposed in the DG literature that can be roughly classified into three lines: learning the domain invariant representation\cite{tzeng2014deep,ghifary2015domain,muandet2013domain,motiian2017unified,wang2018learning}, meta-learning techniques\cite{sankaranarayanan2023meta,balaji2018metareg,dou2019domain,li2018learning} and data perturbation based methods\cite{volpi2018generalizing,qiao2020learning,zhou2020learning}. Our work is most relevant to the last line.

 \textbf{Data perturbation}:
 Data perturbation in the input space can create diverse images to alleviate the spurious correlations\cite{sagawa2019distributionally} and improve the model generalization. Volpi et al.\cite{volpi2018generalizing} proposed an adversarial data augmentation and learned an ensemble model for stable training. Bai et al.\cite{bai2021decaug} decomposed feature representation and semantic augmentation approach for OoD generalization.
 Qiao et al.\cite{qiao2020learning} extended it to create “fictitious” populations with large domain transportation. Zhou et al.\cite{zhou2020learning} employed a data generator to synthesize data from pseudo-novel domains to augment the source domains. Different from these methods, we propose a latent space feature perturbation instead of perturbing raw data in the input space and require no domain labels or any generative-based models.

\textbf{Feature perturbation}:
Unlike most data perturbation methods that adopt transformations in the input space, some approaches perturb features in the latent space. Li et al.\cite{li2021simple} show that even perturbing the feature embedding with Gaussian noise during training leads to a comparable performance. Manifold Mixup \cite{verma2019manifold} adopts linear interpolation from image level to feature level. Recent works show that linear interpolation on feature statistics of two instances \cite{zhou2020domain} can synthesize samples to improve model generalization. Nuriel et al.\cite{nuriel2021permuted} randomly swaps statistics of different samples from the same batch. \cite{li2021uncertainty} extends it and leverages the uncertainty associated with feature perturbations. These methods are based on a fixed perturbation strategy and lack a constraint to preserve semantics. In our work, the LDP modules can generate learnable perturbations to enlarge the domain transportation while the CCFP framework can explicitly preserve the semantic consistency.


\section{Method}

\subsection{General Formulation}
We formulate domain generalization in a classification setting from the input features $x \in \mathcal{X}$ to the predicting labels $y \in \mathcal{Y}$. Given a model family $\Theta$ and training data drawn from some distribution. The goal is to find a model $\theta \in \Theta $ that generalizes well to unseen target distribution $P_{tar}$. DG can be formulated as the following problem:
\begin{equation}\label{1}
\min_{\theta \in \Theta}\mathbb{E}_{(x,y)\sim P_{tar}}[\ell(\theta;(x,y))]
\end{equation}
where $\mathbb{E}[\cdot]$ is the expectation, $\ell(\cdot,\cdot)$ is the loss function.

The challenge for DG is that the target domain distribution $P_{tar}$ is not available. An alternative approach to solve Eq.(\ref{1}) is to merge all the data from source domains and learn the model by minimizing the training error across the pooled data. This is known as the Empirical Risk Minimization (ERM) principle:
\begin{equation}
\hat\theta_{ERM}:=\min_{\theta \in \Theta}\mathbb{E}_{(x,y)\sim P_{src}}[\ell(\theta;(x,y))],
\end{equation}
where $P_{src}$ is the empirical distribution over the training data. Since the ERM-based methods provably lack robustness on OOD scenarios\cite{nagarajan2020understanding,eastwood2022probable}, a number of work\cite{lee2018minimax,heinze2021conditional,robey2021model,sagawa2019distributionally} formulated DG as a worst-case problem leveraging distributionally robust optimization and adversarial training:
\begin{equation}\label{2}
\hat\theta_{worst-case}:=\min_{\theta \in \Theta}\sup_{P:D(P,P_{src}) \leq \rho}\mathbb{E}_{P}[\ell(\theta;(x,y))]
\end{equation}
Here $D(\cdot ,\cdot)$ is a distance metric on the space of probability distributions. The solution to Eq.(\ref{2}) aims to achieve a good performance against the domain shifts while the fictitious target distributions $P$ are distance $\rho$ away from the source domain distribution $P_{src}$.  To solve Eq.(\ref{2}), previous work expects to create fictitious distributions $P$ by perturbing training samples in the input space or using generative-based models and updating the model with respect to these fictitious worst-case target distributions.

However, perturbing samples in the image level may introduce class distortions detrimental to model training which may cause the performance decline. Moreover, the offline two-stage training procedure requires significant computational resources since training the generative model and using it to obtain additional samples are both challenging tasks\cite{wang2019implicit,lu2022action}. 

In this regard, we propose an online one-stage $cross$ $contrasting$ $feature$ $perturbation$ (CCFP) framework (Sec.\ref{3.2}) to obtain perturbed representation distribution $P^l$ with learnable feature statistics (Sec.\ref{3.3}) in latent space without using any generative-based model. Further, to preserve the semantic discriminative information of the perturbed features, we utilize an explicit semantic constraint to encourage the model to predict consistent semantic representations. As it is demanded to determine the source domain distribution and the fictitious target domain distribution according to Eq.\ref{2}, we utilize a dual stream network as it is illustrated in Figure \ref{main}.

It is noteworthy that the distance metric is essential to the worst-case DG problem since it is used to measure the dissimilarity between the source domain distribution and the fictitious target domain distribution. The ideal goal of the metric is to create a fictitious target distribution with a large domain discrepancy from the source distribution as well as retain semantic discriminative information. Previous work directly boosts the dissimilarity in the high-level semantic space\cite{volpi2018generalizing} (usually the output of the last layer), thus failing to preserve the semantic discriminative information. To satisfy the goal, we propose a domain-aware Gram-matrices-based metric to boost the dissimilarity in the whole latent space except for the high-level semantic space (Sec.\ref{3.4}). Further, to better preserve the semantic discriminative information, we utilize a regularization loss to explicitly constrain the semantic consistency between the source domain and the fictitious target domain in Sec.\ref{3.5}.

\subsection{Cross Contrasting Feature Perturbation Framework (CCFP)}\label{3.2}

 Since our goal is to simulate the realistic domain shifts in the latent space. To determine the source domain distribution and create the fictitious target domain distribution, we employ two sub-networks to extract features from the same images. As illustrated in Figure \ref{main}, one is used to extract the original features which represent the online estimation of source distribution in latent space. The other is used to generate perturbed features which represent the fictitious target distribution by using our learnable domain perturbation modules in the intermediate layers.

From the practical perspective, it is intractable to select an appropriate magnitude of the domain shift $\rho$. In this regard, we consider Eq.\ref{2} as the following  Lagrangian relaxation with penalty parameter $\gamma$:
\begin{equation}\label{lagr}
\hat\theta:=\min_{\theta \in \Theta}\sup_{P^l}\{\mathbb{E}_{P^l}[\ell(\theta;(x,y))]-\gamma D(P^l, P_{src}^l)\}
\end{equation}
Here $P_{src}^l$ is the source domain distribution in the latent space, and the $P^l$ is the fictitious target distribution in the latent space.
 Taking the dual reformulation Eq.\ref{lagr}, we can obtain an $min$-$max$ optimization objective that maximizes the domain discrepancy between the source distribution and the fictitious target distribution while minimizing the target risk.

During the $min$-$max$ optimization, each iteration can be divided into two steps. For the maximization step, a batch of images will be fed into both sub-networks and are used to calculate the domain discrepancy loss, denoted by $\mathcal{L}_{dis}$ detailed in Eq.\ref{6}, only the parameters of LDP blocks will be updated at this stage. For the minimization step, the same batch of images will be fed into the model again and are used to calculate the classification loss (cross-entropy loss) and semantic consistency loss, denoted by $\mathcal{L}_{sem}$ detailed in Eq.\ref{sem}, all the parameters of two sub-networks will be updated at this stage.

\subsection{Learnable Domain Perturbation Module (LDP)}\label{3.3}
The key point of our CCFP framework is how to create domain-aware feature perturbation. As perturbing parameters for an affine transformation of intermediate features after normalization can change their characteristics which primarily refer to domain-specific information but are less relevant to category-related information\cite{dumoulin2016learned,huang2017arbitrary}, Huang et al.\cite{huang2017arbitrary} propose the adaptive instance normalization (AdaIN), which replaces the feature statistics of the input features $x$ with the feature statistics of a style image's features $x_{s}$ to achieve style transfer. Let $x \in \mathbb{R}^{B\times C\times H\times W}$ be a batch of features, the AdaIN can be formulated as:
\begin{equation}
AdaIN(x)=\sigma(x_{s}) \frac{x-\mu (x)}{\sigma (x)} + \mu(x_{s})
\end{equation}
\noindent where $\mu (x) \in \mathbb{R}^{B\times C}$ and $\sigma (x) \in \mathbb{R}^{B\times C}$ are the mean and standard deviation respectively.
However, in DG scenarios, the feature statistics of target domain images are not available. Previous work\cite{zhou2020domain, li2021uncertainty} utilizes linear interpolation or uncertainty modeling to diversify the feature statistics, but both of them limit the domain transportation from synthesized features to original features. To address this, we design a learnable domain perturbation (LDP) module (The red box in Figure \ref{main}) to generate perturbed intermediate features:
\begin{equation}\label{LDP}
LDP(x)=(\sigma(x) + \gamma) \frac{x-\mu (x)}{\sigma (x)} + \mu(x) + \beta
\end{equation}
\noindent Here we only add learnable parameters $\gamma$ and $\beta$ to the features' original scaling $\sigma(x)$ and shifting $\mu(x)$ statistics. Different from prior works based on a fixed perturbation strategy, the LDP module can enlarge the domain discrepancy between the original and the perturbed features.

\subsection{Gram-based Domain Discrepancy Metric}\label{3.4}
The worst-case optimization objective for DG is to guarantee model performance against fictitious target distribution within a certain distance from the source distribution. Considering the essential desideratum of DG that enables the model to generalize well to the unseen domain, the ideal distance metric is domain-specific and class-discriminative agnostic. Inspired by the well-known observation\cite{zhang2022contrastive} that the shallow layers learn low-level features which are task-irrelevant, we build an effective domain discrepancy metric applied to the shallow layers. Specifically, Gatys et al.\cite{gatys2015texture} shows that Gram matrices can encode stylistic attributes like textures and patterns that are less relevant to task objectives but can be used to depict the individual domain information. Therefore, we utilize the Gram-matrices-based metric to depict the domain discrepancy.

Specifically, we denote the network as the following:
\begin{equation}
c(x) = g \circ f^n \circ f^{n-1} \circ \cdots \circ f^1(x)
\end{equation}
Here $g$ is the classifier, $f=f^n \circ f^{n-1} \circ \cdots \circ f^1(x)$ is the feature extractor, and $n$ denotes the number of shallow layers. In our method, we use a set of Gram matrices $\{G^1, G^2, \cdots, G^K \}$ from a set of shallow layers $\{f^1, \cdots, f^K\}$ in the network to describe the domain-specific characteristics. The domain discrepancy loss can be formulated as:
\begin{equation}\label{6}
\mathcal{L}_{dis} = -\sum_{i=1}^K||G(f_o^i(\mathbf{x})) - G(f_p^i(\mathbf{x}))||_F 
\end{equation}

\noindent where $f_o$ and $f_p$ are two feature extractors (original and perturbed) respectively. $K$ is the number of shallow layers to calculate the loss $\mathcal{L}_{dis}$, $G(\cdot)$ is the Gram matrix, and the $||\cdot||_F$ denotes the Frobenius norm.

\subsection{Explicit Semantic Consistency Constraint}\label{3.5}
To better preserve the semantic consistency, we minimize the L2-norm between the final classifier predictions of two sub-networks. The semantic consistency loss can be formulated as:
\begin{equation}\label{sem}
L_{sem} = ||g_o(f_o(\mathbf{x})) - g_p(f_p(\mathbf{x}))||_2^2
\end{equation}
Here $g_o$ and $g_p$ are two classifiers (original and perturbed respectively). The final loss is given by:
\begin{equation}\label{eq_final}
\mathcal{L}_{final} = \mathcal{L}_{cls_1} + \mathcal{L}_{cls_2} + \lambda_{dis} \mathcal{L}_{dis} + \lambda_{sem} \mathcal{L}_{sem}
\end{equation}
\noindent The $\lambda_{dis}$ and the $\lambda_{sem}$ are used to control the strength of the domain discrepancy loss $\mathcal{L}_{dis}$ and the semantic consistency loss $\mathcal{L}_{sem}$.

The optimization algorithm is designed in Algorithm \ref{alg1}.
During the inference, we only use the sub-network (the top network in the Figure \ref{main}) which is learned from the perturbed features to predict the final results, as the diverse latent features help mitigate the domain shift. Additionally, the LDP modules will also be used to prevent variations in the normalization statistics that could otherwise cause model collapse. Although the statistics shift can be alleviated by randomly applying LDP during training, the LDP modules can also be used as a test-time augmentation technology to boost the performance, we will discuss it in the Appendix.

\begin{algorithm}[t]\label{alg1}
	\renewcommand{\algorithmicrequire}{\textbf{Input:}}
	\renewcommand{\algorithmicensure}{\textbf{Output:}}
        \setstretch{1.1}
	\caption{\textbf{:} Cross Contrasting Feature Perturbation}
	\begin{algorithmic}
		\STATE \textbf{Input:}$\mathcal{S}_{train}=\{( \mathbf{x}_i,y_i)\}^{n}_{i=1}$, batch size $B$, learning rate $\eta$, Adam optimizer, initial $\lambda_{dis}, \lambda_{sem}$
		\STATE \textbf{Initial:} Parameters of CCFP $i.e.$ parameters $\theta_0$, $\theta_1$, $\phi_0$, $\phi_1$, $(\gamma_k, \beta_k; k=1 \cdots K)$ for feature extractor $f_o$, $f_p$, classifier $g_o, g_p$ and LDP modules $P^1, P^2 \cdots P^K$ (K is defined in Eq.\ref{6}).
		\REPEAT
            \STATE \textbf{Minimization Stage:}
            \FOR{$i=1,\cdots,B$}{
                \STATE $\mathcal{L}_{cls_1}^i=\ell(g_o(f_o(\mathbf{x}_i)),y_i)$
                \STATE $\mathcal{L}_{cls_2}^i=\ell(g_p(f_p(\mathbf{x}_i)),y_i)$
                \STATE $\mathcal{L}_{sem}^i=\lambda_{sem} ||f_o(\mathbf{x}_i)-f_p(\mathbf{x}_i)||^2_2$
            }\ENDFOR
            \begin{small}
            \STATE $\theta_0,\phi_0 \leftarrow $ Adam$(\frac{1}{B}\sum_{i=1}^{B}\mathcal{L}_{cls_1}^i+\mathcal{L}_{sem}^i,\theta_0,\phi_0,\eta)$
            \end{small}
            \begin{small}
            $\theta_1,\phi_1,\gamma_k,\beta_k \leftarrow $ Adam$(\frac{1}{B}$
            $\sum_{i=1}^B\mathcal{L}_{cls_2}^i+\mathcal{L}_{sem}^i,\theta_1,\phi_1,$ \\
            \hspace{3.2cm}$\gamma_k,\beta_k,\eta$)
            \end{small}
            \STATE \textbf{Maximization Stage:}
            \FOR{$i=1,\cdots,B$}{
                \STATE\hspace{-0.2cm}$\mathcal{L}_{dis}^i=\lambda_{dis} \sum_{k=1}^K||G(f_o^k(\mathbf{x}_i)) - G(f_p^k(P^k(\mathbf{x}_i)))||_F$
            }\ENDFOR
            \STATE $\gamma_k,\beta_k \leftarrow$ Adam$(\frac{1}{B}\sum_{i=1}^B \mathcal{L}_{spe}^j,\gamma_k,\beta_k,\eta)$
		\UNTIL $\theta_0$, $\theta_1$, $\phi_0$, $\phi_1$ are converged
	\end{algorithmic}  
\end{algorithm}

\begin{table*}[t]
  \centering
  \begin{tabular}{c | c c c c c c c | c}
    \toprule
    \textbf{Algorithm} & \textbf{CMNIST} & \textbf{RMNIST} & \textbf{VLCS} & \textbf{PACS} & \textbf{OfficeHome} & \textbf{TerraInc} & \textbf{DomainNet} & \textbf{Avg} \\
    \midrule
    ERM\cite{vapnik1991principles} & 51.5 \footnotesize$\pm$ 0.1 & 98.0 \footnotesize$\pm$ 0.0 & 77.5 \footnotesize$\pm$ 0.4 & 85.5 \footnotesize$\pm$ 0.2 & 66.5 \footnotesize$\pm$ 0.3 & 46.1 \footnotesize$\pm$ 1.8 & 40.9 \footnotesize$\pm$ 0.1 & 66.6\\
    IRM\cite{arjovsky2020invariant} & 52.0 \footnotesize$\pm$ 0.1 & 97.7 \footnotesize$\pm$ 0.1 & 78.5 \footnotesize$\pm$ 0.5 & 83.5 \footnotesize$\pm$ 0.8 & 64.3 \footnotesize$\pm$ 2.2 & 47.6 \footnotesize$\pm$ 0.8 & 33.9 \footnotesize$\pm$ 2.8 & 65.4\\
    GroupDRO\cite{sagawa2019distributionally} & 52.1 \footnotesize$\pm$ 0.0 & 98.0 \footnotesize$\pm$ 0.0 & 76.7 \footnotesize$\pm$ 0.6 & 84.4 \footnotesize$\pm$ 0.8 & 66.0 \footnotesize$\pm$ 0.7 & 43.2 \footnotesize$\pm$ 1.1 & 33.3 \footnotesize$\pm$ 0.2 & 64.8 \\
    Mixup\cite{yan2020improve} & 52.1 \footnotesize$\pm$ 0.2 & 98.0 \footnotesize$\pm$ 0.1 & 77.4 \footnotesize$\pm$ 0.6 & 84.6 \footnotesize$\pm$ 0.6 & 68.1 \footnotesize$\pm$ 0.3 & 47.9 \footnotesize$\pm$ 0.8 & 39.2 \footnotesize$\pm$ 0.1 & 66.7\\
    MLDG\cite{li2018learning}& 51.5 \footnotesize$\pm$ 0.1 & 97.9 \footnotesize$\pm$ 0.0 & 77.2 \footnotesize$\pm$ 0.4 & 84.9 \footnotesize$\pm$ 1.0 & 66.8 \footnotesize$\pm$ 0.6 & 47.7 \footnotesize$\pm$ 0.9 & 41.2 \footnotesize$\pm$ 0.1 & 66.7\\
    CORAL\cite{sun2016deep} & 51.5 \footnotesize$\pm$ 0.1 & 98.0 \footnotesize$\pm$ 0.1 & 78.8 \footnotesize$\pm$ 0.6 & 86.2 \footnotesize$\pm$ 0.3 & 68.7 \footnotesize$\pm$ 0.3 & 47.6 \footnotesize$\pm$ 1.0 & 41.5 \footnotesize$\pm$ 0.1 & 67.5\\
    MMD\cite{li2018domain} & 51.5 \footnotesize$\pm$ 0.2 & 97.9 \footnotesize$\pm$ 0.0 & 77.5 \footnotesize$\pm$ 0.9 & 84.6 \footnotesize$\pm$ 0.5 & 66.3 \footnotesize$\pm$ 0.1 & 42.2 \footnotesize$\pm$ 1.6 & 23.4 \footnotesize$\pm$ 9.5 & 63.3\\
    DANN\cite{ganin2016domain} & 51.5 \footnotesize$\pm$ 0.2 & 97.8 \footnotesize$\pm$ 0.1 & 78.6 \footnotesize$\pm$ 0.4 & 83.6 \footnotesize$\pm$ 0.4 & 65.9 \footnotesize$\pm$ 0.6 & 46.7 \footnotesize$\pm$ 0.5 & 38.3 \footnotesize$\pm$ 0.1 & 66.1\\
    CDANN\cite{li2018domain} & 51.7 \footnotesize$\pm$ 0.1 & 97.9 \footnotesize$\pm$ 0.1 & 77.5 \footnotesize$\pm$ 0.1 & 82.6 \footnotesize$\pm$ 0.9 & 65.8 \footnotesize$\pm$ 1.3 & 45.8 \footnotesize$\pm$ 1.6 & 38.3 \footnotesize$\pm$ 0.3 & 65.6\\
    MTL\cite{blanchard2021domain} & 51.4 \footnotesize$\pm$ 0.1 & 97.9 \footnotesize$\pm$ 0.0 & 77.2 \footnotesize$\pm$ 0.4 & 84.6 \footnotesize$\pm$ 0.5 & 66.4 \footnotesize$\pm$ 0.5 & 45.6 \footnotesize$\pm$ 1.2 & 40.6 \footnotesize$\pm$ 0.1 & 66.2\\
    SagNet\cite{nam2021reducing} & 51.7 \footnotesize$\pm$ 0.0 & 98.0 \footnotesize$\pm$ 0.0 & 77.8 \footnotesize$\pm$ 0.5 & 86.3 \footnotesize$\pm$ 0.2 & 68.1 \footnotesize$\pm$ 0.1 & \textbf{48.6} \footnotesize$\pm$ 1.0 & 40.3 \footnotesize$\pm$ 0.1 & 67.2\\
    ARM\cite{zhang2020adaptive} & \textbf{56.2} \footnotesize$\pm$ 0.2 & \textbf{98.2} \footnotesize$\pm$ 0.1 & 77.6 \footnotesize$\pm$ 0.3 & 85.1 \footnotesize$\pm$ 0.4 & 64.8 \footnotesize$\pm$ 0.3 & 45.5 \footnotesize$\pm$ 0.3 & 35.5 \footnotesize$\pm$ 0.2 & 66.1\\
    V-REx\cite{krueger2021out} & 51.8 \footnotesize$\pm$ 0.1 & 97.9 \footnotesize$\pm$ 0.1 & 78.3 \footnotesize$\pm$ 0.2 & 84.9 \footnotesize$\pm$ 0.6 & 66.4 \footnotesize$\pm$ 0.6 & 46.4 \footnotesize$\pm$ 0.6 & 33.6 \footnotesize$\pm$ 2.9 & 65.6\\
    RSC\cite{huang2020self} & 51.7 \footnotesize$\pm$ 0.2 & 97.6 \footnotesize$\pm$ 0.1 & 77.1 \footnotesize$\pm$ 0.5 & 85.2 \footnotesize$\pm$ 0.9 & 65.5 \footnotesize$\pm$ 0.9 & 46.6 \footnotesize$\pm$ 1.0 & 38.9 \footnotesize$\pm$ 0.5 & 66.1\\
    \midrule
    AND-mask\cite{kim2021selfreg} & 51.3 \footnotesize$\pm$ 0.2 & 97.6 \footnotesize$\pm$ 0.1 & 78.1 \footnotesize$\pm$ 0.9 & 84.4 \footnotesize$\pm$ 0.9 & 65.6 \footnotesize$\pm$ 0.4 & 44.6 \footnotesize$\pm$ 0.3 & 37.2 \footnotesize$\pm$ 0.6 & 65.5\\
    SAND-mask\cite{kim2021selfreg} & 51.8 \footnotesize$\pm$ 0.2 & 97.4 \footnotesize$\pm$ 0.1 & 77.4 \footnotesize$\pm$ 0.2 & 84.6 \footnotesize$\pm$ 0.9 & 65.8 \footnotesize$\pm$ 0.4 & 42.9 \footnotesize$\pm$ 1.7 & 32.1 \footnotesize$\pm$ 0.6 & 64.6\\
    Fish\cite{shi2021gradient} & 51.6 \footnotesize$\pm$ 0.1 & 98.0 \footnotesize$\pm$ 0.0 & 77.8 \footnotesize$\pm$ 0.3 & 85.5 \footnotesize$\pm$ 0.3 & 68.6 \footnotesize$\pm$ 0.4 & 45.1 \footnotesize$\pm$ 1.3 & \textbf{42.7} \footnotesize$\pm$ 0.2 & 67.1\\
    Fishr\cite{rame2022fishr} & 52.0 \footnotesize$\pm$ 0.2 & 97.8 \footnotesize$\pm$ 0.0 & 77.8 \footnotesize$\pm$ 0.1 & 85.5 \footnotesize$\pm$ 0.4 & 67.8 \footnotesize$\pm$ 0.1 & 47.4 \footnotesize$\pm$ 1.6 & 41.7 \footnotesize$\pm$ 0.0 & 67.1\\
    \midrule
    CCFP (ours) & 51.9 \footnotesize$\pm$ 0.1 & 97.8 \footnotesize$\pm$ 0.1 & \textbf{78.9} \footnotesize$\pm$ 0.3 & \textbf{86.6} \footnotesize$\pm$ 0.2 & \textbf{68.9} \footnotesize$\pm$ 0.1 & \textbf{48.6} \footnotesize$\pm$ 0.4 & 41.2 \footnotesize$\pm$ 0.0 & \textbf{67.7} \\
    \bottomrule
  \end{tabular}
  \caption{DomainBed with Training-domain model selection. We highlighted the best results using \textbf{bold} font.}
  \label{result}
\end{table*}

\section{Experiments}

\subsection{DomainBed Benchmark}\label{4.1}
We conduct comprehensive experiments on the DomainBed benchmark\cite{gulrajani2020search}. DomainBed includes seven multi-domain image classification tasks: Colored MNIST\cite{arjovsky2020invariant}, Rotated MNIST\cite{ghifary2015domain}, PACS\cite{li2017deeper}, VLCS\cite{fang2013unbiased}, Office-Home\cite{venkateswara2017deep}, Terra Incognita\cite{beery2018recognition}, and DomainNet\cite{peng2019moment}.

\textbf{Colored MNIST}\cite{arjovsky2020invariant} is a variant of MNIST consisting of 70,000 examples of dimension (2, 28, 28) and 2 classes. The dataset contains a disjoint set of colored digits where domain $d$ $\in$ \{90\%, 80\%, 10\%\} is the correlation strength between color and label across domains. \textbf{Rotated MNIST}\cite{ghifary2015domain} is a variant of MNIST consisting of 70,000 examples of dimension (1, 28, 28) and 10 classes. The dataset contains digits rotated by $d$ degrees where domain $d$ $\in$ \{0, 15, 30, 45, 60, 75\}. \textbf{PACS}\cite{li2017deeper} includes domains $d$ $\in$ \{art, cartoons, photos, sketches\} with 9,991 examples of dimension (3, 224, 224) and 7 classes. \textbf{VLCS}\cite{fang2013unbiased} includes domains $d$ $\in$\{Caltech101, LabelMe, SUN09, VOC2007\} with 10,729 examples of dimension (3, 224, 224) and 5 classes. \textbf{Office-Home}\cite{venkateswara2017deep} includes domains $d$ $\in$ \{atr, clipart, product, real\} with 15,588 examples of dimension (3, 224, 224) and 65 classes. \textbf{Terra Incognita}\cite{beery2018recognition} contains photographs of wild animals taken by camera traps at locations $d$ $\in$ \{L100, L38, L43, L46\} with 24,788 examples of dimension (3, 224, 224) and 10 classes. \textbf{DomainNet}\cite{peng2019moment} includes domains $d$ $\in$ \{clipart, infograph, painting, quickdraw, real, sketch\} with 586,575 examples of dimension (3, 224, 224) and 345 classes.

For a fair comparison, the DomainBed benchmark\cite{gulrajani2020search} presents an evaluation protocol about dataset splits, model selection on the validation set, and hyperparameter (HP) search, which is detailed below.

\textbf{Dataset splits.}
The data from source domains are split into training subsets (80\%) and validation subsets (20\%) (used on Training-domain validation set model selection). The data from the target domain are split into testing subsets (80\%) and validation subsets (20\%) (used on Test-domain validation set model selection). We repeat the entire experiment three times using different seeds and report the mean and standard error over all the repetitions.

\textbf{Model selection methods.}
There are three model selection methods in \cite{gulrajani2020search}. ($\romannumeral1$) Training-domain validation set. ($\romannumeral2$) Leave-one-out cross-validation. ($\romannumeral3$) Test-domain validation set (oracle). 
We choose the Training-domain model selection that assumes the training and test examples follow similar distributions. The best-performing model in the validation set is selected as the final model, and its test domain performance is reported as the final performance. The results of the oracle model selection are shown in Appendix.

\textbf{Model architectures.}
Following DomainBed, we use Conv-Net (detail in Appendix D.1 in \cite{gulrajani2020search}) as the backbone for Colored MNIST and Rotated MNIST and use ResNet-50\cite{he2016deep} for the rest datasets. For the classifier, we only use one linear layer. We insert the LDP modules after the first Conv, Max Pooling, and 1,2,3-th ConvBlock, and we further perform an ablation study for the effects of different inserted positions. When using Conv-Net as our backbone, we insert the LDP modules at the position after the first three Batch Normalization layers.

\textbf{Hyperparameter (HP) search.} We run a random search of 20 trials over the hyperparameter distribution given by DomainBed. Our CCFP relies on two additional hyperparameters $\lambda_{spe}$ and $\lambda_{sem}$, and we set the range of search such as $[0.1, 10]$ for both of them, more details about the range of hyperparameter search will be discussed in Appendix.

\textbf{Implementation details.}
We implement our algorithm using the codebase of DomainBed in PyTorch, using ResNet-50 pre-trained on the ImageNet\cite{deng2009imagenet} and fine-tuning on each dataset. Note that our evaluation setting follows the standard evaluation protocol given by DomainBed\cite{gulrajani2020search}.

\subsection{Results}

\textbf{Comparison with domain generalization methods on Domainbed benchmark.}
Comprehensive experiments show that CCFP achieves significant performance gain against previous methods on most of the benchmark datasets and obtains comparable performance on three of seven datasets. Table \ref{result} summarizes the results on DomainBed using the Training-domain model selection method. Our CCFP outperforms all previous approaches on the averaged result. 

To further validate the generalization of CCFP, we conduct experiments under another commonly used baseline SWAD\cite{cha2021swad} as our backbone, which is a unique model selection mechanism. For a fair comparison, we only summarize the methods based on SWAD. The performance comparison with other existing approaches that adopted SWAD is provided in Tables \ref{OfficeHome}-\ref{Terra}. Our CCFP achieves significant performance gain in all experiments against previous best results.

\begin{table}[ht]
  \centering
  \begin{tabular}{c c c c c c}
    \toprule
    \textbf{Algorithm} & \textbf{A} & \textbf{C} & \textbf{P} & \textbf{R} & \textbf{Avg.} \\
    \midrule
    SWAD\cite{cha2021swad} & 66.1 & 57.7 & 78.4 & 80.2 & 70.6\\
    PCL\cite{yao2022pcl} & 67.3 & \textbf{59.9} & 78.7 & 80.7 & 71.6 \\
    \midrule
    CCFP (ours) & \textbf{68.0} & 58.6 & \textbf{79.7} & \textbf{81.9} & \textbf{72.1}\\
    \bottomrule
  \end{tabular}
  \caption{Comparison with SWAD-based state-of-the-art methods on OfficeHome benchmark. A: art, C: clipart, P: product, R: real, Avg.: average.}
  \label{OfficeHome}
\end{table}

\begin{table}[ht]
  \centering
  \begin{tabular}{c c c c c c}
    \toprule
    \textbf{Algorithm} & \textbf{C} & \textbf{L} & \textbf{S} & \textbf{V} & \textbf{Avg.} \\
    \midrule
    SWAD\cite{cha2021swad} & 98.8 & 63.3 & \textbf{75.3} & 79.2 & 79.1\\
    PCL\cite{yao2022pcl} & \textbf{99.0} & 63.6 & 73.8 & 75.6 & 78.0 \\
    \midrule
    CCFP (ours) & 98.9 & \textbf{64.1} & 74.9 & \textbf{79.9} & \textbf{79.4}\\
    \bottomrule
  \end{tabular}
  \caption{Comparison with SWAD-based state-of-the-art methods on VLCS benchmark. C: Caltech101, L: LabelMe, S: SUN09, V: VOC2007, Avg.: average.}
  \label{VLCS}
\end{table}

\begin{table}[ht]
  \centering
  \begin{tabular}{c c c c c c}
    \toprule
    \textbf{Algorithm} & \textbf{L100} & \textbf{L38} & \textbf{L43} & \textbf{L46} & \textbf{Avg.} \\
    \midrule
    SWAD\cite{cha2021swad} & 55.4 & 44.9 & 59.7 & 39.9 & 50.0\\
    PCL\cite{yao2022pcl} & 58.7 & 46.3 & 60.0 & 43.6 & 52.1 \\
    \midrule
    CCFP (ours) & \textbf{59.9} & \textbf{47.6} & \textbf{60.8} & \textbf{43.8} & \textbf{53.0}\\
    \bottomrule
  \end{tabular}
  \caption{Comparison with SWAD-based state-of-the-art methods on TerraIncognita benchmark. L100: Location 100, L38: Location 38, L43: Location 43, L46: Location 46, Avg.: average.}
  \label{Terra}
\end{table}

\textbf{Comparison with previous feature perturbation methods.}
To reveal the performance gain by using learnable parameters to perturb feature statistics, we conduct experiments to compare with two previous features perturbation methods Mixstyle\cite{zhou2020domain} and DSU\cite{li2021uncertainty}. Since both of them use their own experiment settings. For a fair comparison, we rerun their results on the DomainBed experiment benchmark. Table \ref{DSU} shows that our CCFP achieves a substantial improvement in performance compared to the previous feature perturbation methods more experiment results are shown in Appendix.

\begin{table}[ht]
  \centering
  \begin{tabular}{c c c c c c}
    \toprule
    \textbf{Algorithm} & \textbf{A} & \textbf{C} & \textbf{P} & \textbf{S} & \textbf{Avg.} \\
    \midrule
    ERM & 81.6 & 78.7 & 95.5 & 78.7 & 83.6\\
    Mixstyle\cite{zhou2020domain} & 84.0 & 79.9 & 94.3 & \textbf{81.6} & 84.9 \\
    DSU\cite{li2021uncertainty} & 81.9 & 79.6 & 95.0 & 79.6 & 84.1 \\
    \midrule
    CCFP (ours) & \textbf{87.5} & \textbf{81.3} & \textbf{96.4} & 81.4 & \textbf{86.6}\\
    \bottomrule
  \end{tabular}
  \caption{Comparison with previous feature perturbation methods on PACS benchmark. Comparison with SWAD-based state-of-the-art methods on PACS benchmark.}
  \label{DSU}
\end{table}

\section{Ablation Study}
\textbf{Effects of the explicit semantic regularization.} To validate the effectiveness of the semantic regularization, we conduct experiments without using the semantic consistency loss in Eq.\ref{eq_final}. Table \ref{semantic} shows that semantic regularization can achieve performance gain on most target domains and the average accuracy. In particular, we can find that without semantic regularization, our method can still significantly outperform ERM.

\begin{table}[ht]
  \centering
  \begin{tabular}{c c c c c c}
    \toprule
    \textbf{Algorithm} & \textbf{A} & \textbf{C} & \textbf{P} & \textbf{S} & \textbf{Avg.} \\
    \midrule
    ERM & 81.6 & 78.7 & 95.5 & 78.7 & 83.6\\
    CCFP(w/o) $L_{sem}$ & 83.6 & \textbf{83.9} & \textbf{96.4} & 80.3 & 86.0 \\
    \midrule
    CCFP (ours) & \textbf{87.5} & 81.3 & \textbf{96.4} & \textbf{81.4} & \textbf{86.6}\\
    \bottomrule
  \end{tabular}
  \caption{Comparison with result without using $L_{sem}$ on PACS benchmark.}
  \label{semantic}
\end{table}

\begin{figure*}[htbp]
\centering
\begin{minipage}[t]{0.24\textwidth}
\centering
\includegraphics[width=1\linewidth]{
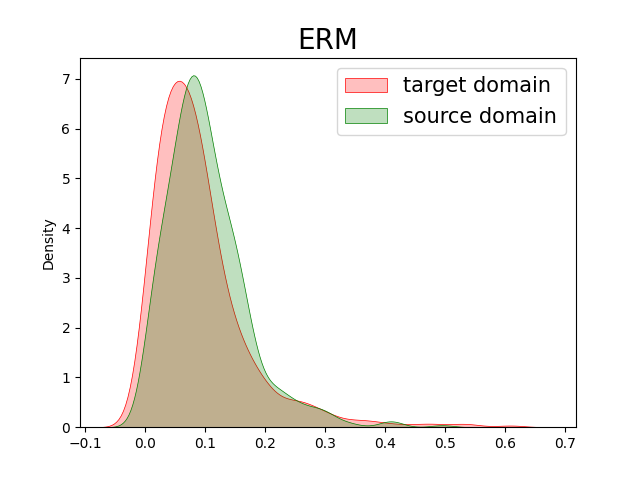}
\end{minipage}
\begin{minipage}[t]{0.24\textwidth}
\centering
\includegraphics[width=1\linewidth]{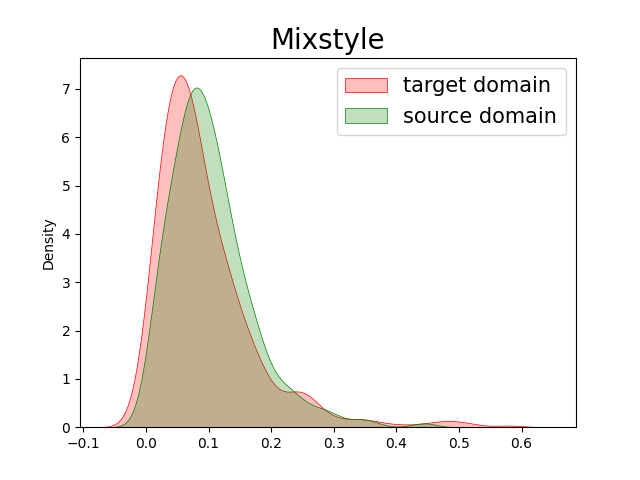}
\end{minipage}
\begin{minipage}[t]{0.24\textwidth}
\centering
\includegraphics[width=1\linewidth]{
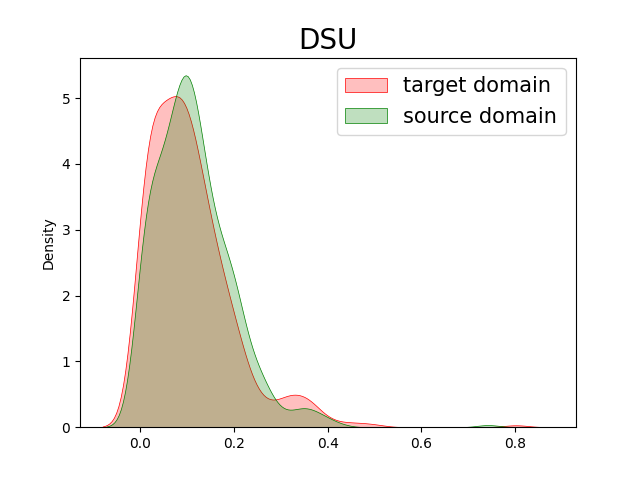}
\end{minipage}
\begin{minipage}[t]{0.24\textwidth}
\centering
\includegraphics[width=1\linewidth]{
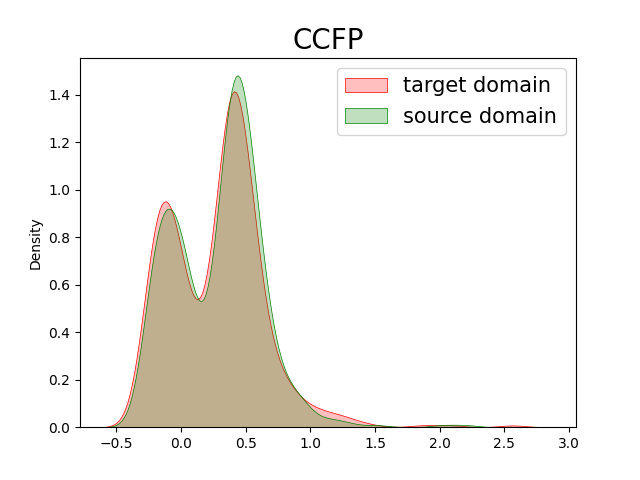}
\end{minipage}
\label{shift0}
\end{figure*}
\begin{figure*}[htbp]
\centering
\begin{minipage}[t]{0.24\textwidth}
\centering
\includegraphics[width=1\linewidth]{
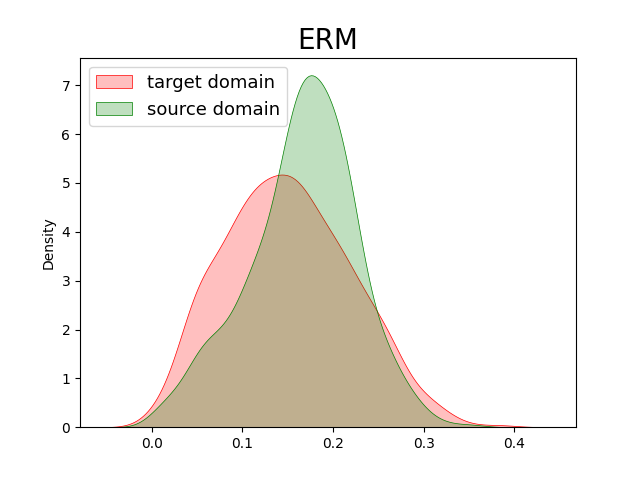}
\end{minipage}
\begin{minipage}[t]{0.24\textwidth}
\centering
\includegraphics[width=1\linewidth]{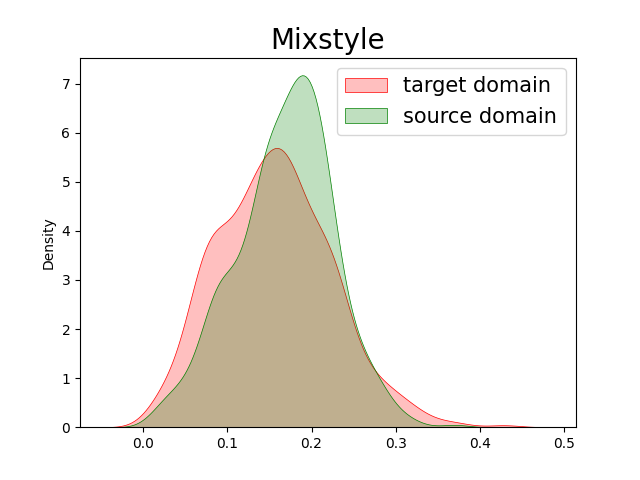}
\end{minipage}
\begin{minipage}[t]{0.24\textwidth}
\centering
\includegraphics[width=1\linewidth]{
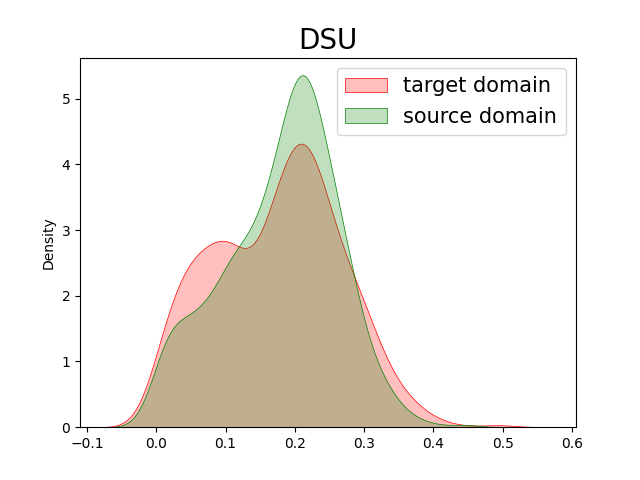}
\end{minipage}
\begin{minipage}[t]{0.24\textwidth}
\centering
\includegraphics[width=1\linewidth]{
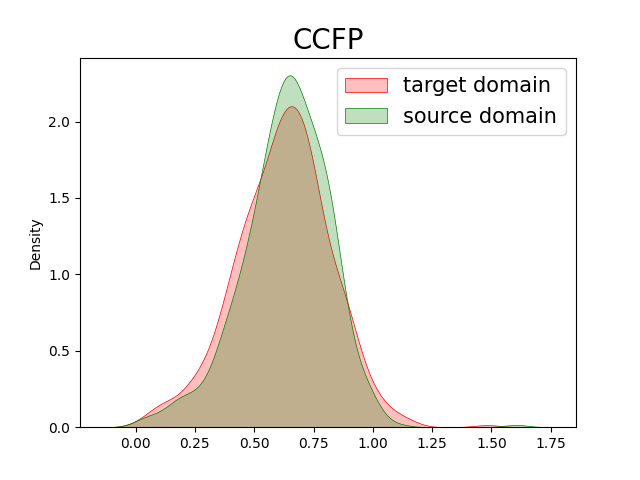}
\end{minipage}
\caption{The visualization of feature statistics at the position 3. The top raw is the mean statistics and the bottom raw is the std statistics. We conduct the experiments on the PACS dataset with ERM, Mixstyle, DSU and our CCFP.}
\label{shift1}
\end{figure*}

To further validate the essential to regularize the semantic consistency after perturbing features in the latent space, we enforce the semantic regularization on previous feature perturbation methods. Note that both Mixstyle and DSU use one single network to generate the perturbed features, which is unable to calculate the $L_{sem}$ in Eq.\ref{eq_final}. To address this, we implement the two methods in our CCFP framework. Similar to our approach, we use one sub-network to extract the original features and use the other sub-network to generate the perturbed features by using Mixstyle and DSU feature perturbation methods. Further, we constrain the consistency between the predictions of the two sub-networks. During the inference, we only use the perturbed sub-network to produce the final predictions which are the same as our approach. Since Mixstyle and DSU are non-parametric, we remove the $L_{dis}$ in Eq.\ref{eq_final} in this experiment. Table \ref{arch} shows that our dual stream architecture and the explicit semantic consistency regularization can achieve a significant performance gain (0.3$\%$ for Mixstyle and 1.4$\%$ for DSU).

\begin{table}[ht]
  \centering
  \begin{tabular}{c c c c c c}
    \toprule
    \textbf{Algorithm} & \textbf{A} & \textbf{C} & \textbf{P} & \textbf{S} & \textbf{Avg.} \\
    \midrule
    Mixstyle\cite{zhou2020domain} & 84.0 & 79.9 & 94.3 & 81.6 & 84.9 \\
    Mixstyle (dual) & 84.6 & 80.3 & \textbf{96.5} & 79.5 & 85.2 \\
    DSU\cite{li2021uncertainty} & 81.9 & 79.6 & 95.0 & 79.6 & 84.1 \\
    DSU (dual) & 86.3 & 79.4 & 94.6 & \textbf{81.7} & 85.5 \\
    \midrule
    CCFP (ours) & \textbf{87.5} & \textbf{81.3} & 96.4 & 81.4 & \textbf{86.6}\\
    \bottomrule
  \end{tabular}
  \caption{Validation of the additional semantic consistency for previous feature perturbation methods on PACS benchmark.}
  \label{arch}
\end{table}

\noindent \textbf{Effects of LDP inserted positions.}
In CCFP, we use a set of Gram matrices of intermediate features from a set of layers $\{f^1, \cdots, f^K\}$ to describe the domain-specific characteristics\cite{zhang2022refining, zhang2022characterizing}. To verify the effects of LDP inserted positions, we name the position of ResNet after the first Conv, Max Pooling, and 1,2,3-th ConvBlock as 1,2,3,4,5 respectively, and the effects of LDP on different inserted positions are evaluated accordingly. We conduct the experiments on dataset PACS and OfficeHome with the default hyperparameters given by DomainBed. Table \ref{abla1} shows that more inserted LDP modules can produce relatively higher classification accuracy. Hence we plug the LDP modules into all five positions for our main experiments. 
\begin{table}[ht]
  \centering
  \begin{tabular}{c c c c c |c}
    \toprule
    \textbf{Positions} & \textbf{1-3} & \textbf{2-4} & \textbf{3-5} & \textbf{1-5} & \textbf{ERM}\\
    \midrule
    PACS & 85.3 & 84.8 & 85.4 & \textbf{86.6} & 83.6\\
    OfficeHome & 68.4 & 68.5 & 68.3 & \textbf{68.9}& 64.5\\
    \bottomrule
  \end{tabular}
  \caption{Effects of different inserted positions on PACS and OfficeHome benchmark.}
  \label{abla1}
\end{table}

\noindent \textbf{Visualization analysis on CCFP.} 
To confirm that our CCFP can alleviate the domain shift phenomena, we conduct experiments on the PACS dataset where we choose art painting as the target domain and the rest as the source domain. We capture the intermediate features at position 4 to study the feature statistic shifts. Figure \ref{shift1} shows the feature statistics distribution from source domains and target domains based on ERM, Mixstyle, DSU, and our CCFP. For a fair comparison, we reproduce the results of ERM, Mixstyle, DSU, and CCFP with the same fixed steps (5,000 steps, which is the same as the default value given by DomainBed) and only consider the final checkpoint. It is shown that our CCFP can obviously mitigate the domain shift between the source and target domain features compared with ERM and surpass Mixstyle and DSU. The result shows that our method can help against the domain shift.

\section{Conclusions}
In this paper, we propose a simple yet efficient cross contrasting feature perturbation framework. Unlike previous works, our method does not use generative-based models or domain labels. Our approach can adaptively generate perturbed features with large domain transportation from the original features while preserving semantic consistency, and encourage the model to predict consistent semantic representation against the domain shift. The experiments show that our method performs better than the previous state-of-the-art on the DomainBed benchmark.

\section{Acknowledgments}
This work is supported by National Key Research and Development Program of China (2021YFF1200800)

{\small
\bibliographystyle{ieee_fullname}
\bibliography{egbib}
}

\end{document}